\documentclass[10pt,twocolumn,letterpaper]{article}

\usepackage{iccv}
\usepackage{times}
\usepackage{epsfig}
\usepackage{graphicx}
\usepackage{amssymb}
\usepackage{cite}
\usepackage{amsmath,amssymb,amsfonts}
\usepackage{algorithmic}
\usepackage{pbox}
\usepackage{textcomp}
\usepackage{xcolor}
\usepackage{booktabs}
\usepackage{multirow}
\usepackage[inline]{enumitem}

\DeclareMathOperator{\EX}{\mathbb{E}}

\usepackage[breaklinks=true,bookmarks=false]{hyperref}
\usepackage{cleveref}

\iccvfinalcopy 

\ificcvfinal\fi

\begin{document}

\title{Conditional Generative Adversarial Networks for Speed Control in Trajectory Simulation}

\author{Sahib Julka\\
University of Passau\\
Germany\\
{\tt\small sahib.julka@uni-passau.de}

\and
Vishal Sowrirajan\\
University of Passau\\
Germany\\
{\tt\small sowrir01@ads.uni-passau.de}
\and
Joerg Schloetterer\\
University of Duisburg-Essen\\
Germany\\
{\tt\small joerg.schloetterer@uni-due.de}
\and
Michael Granitzer\\
University of Passau\\
Germany\\
{\tt\small michael.granitzer@uni-passau.de}
\and
}

\maketitle

\begin{abstract}

Motion behaviour is driven by several factors - goals, presence and actions of neighbouring agents, social relations, physical and social norms, the environment with its variable characteristics, and further. 
Most factors are not directly observable and must be modelled from context. Trajectory prediction, is thus a hard problem, and has seen increasing attention from researchers in the recent years. Prediction of motion, in application, must be realistic, diverse and controllable. In spite of increasing focus on multimodal trajectory generation, most methods still lack means for explicitly controlling different modes of the data generation. Further, most endeavours invest heavily in designing special mechanisms to learn the interactions in latent space. We present Conditional Speed GAN (CSG), that allows controlled generation of diverse and socially acceptable trajectories, based on user controlled speed. During prediction, CSG forecasts future speed from latent space and conditions its generation based on it. CSG is comparable to state-of-the-art GAN methods in terms of the benchmark distance metrics, while being simple and useful for simulation and data augmentation for different contexts such as fast or slow paced environments. Additionally, we compare the effect of different aggregation mechanisms and show that a naive approach of concatenation works comparable to its attention and pooling alternatives.

\end{abstract}

\section{Introduction}

Modelling social interactions and the ability to forecast motion dynamics is pertinent to several application domains such as robot planning systems~\cite{chen2019crowd}, traffic operations~\cite{horni2016multi}, and autonomous vehicles~\cite{rasouli2019autonomous}. However, it remains a challenge due to the subjectivity and variability of interactions in real world scenarios. Trajectory prediction not only needs to be sensitive to several real world constraints, but also involves implicit semantic modelling of an agents mobility patterns, while anticipating the movements of other agents in the scene. 

Recently we have witnessed a shift in perspective from the more deterministic approaches of agent modelling with handcrafted features~\cite{social_force_model_17, discrete_choice_2, wang2007gaussian, handcrafted3,handcrafted2, handcrafted1}, to the latent learning of variable outcomes via complex data-driven deep neural network architectures~\cite{base_paper, SocialWays, SophiePaper, Trajectron++}. State-of-the-art systems are able to generate variable or multimodal predictions that are socially acceptable (adhere to social norms), spatially aware and similar to the semantics in training data. Most systems can sufficiently generate outcomes according to the original distribution, but lack means for controlling different modes of data generation, or to be able to extrapolate to unseen contexts. Consequently, controlled simulation is a challenge. 

Furthermore, most approaches focus on modelling of a single agent i.e., pedestrians~\cite{base_paper} or vehicles~\cite{paden2016survey}, and lack, thus, the  modelling of heterogeneous semantic classes. We propose that these systems need to be \begin{enumerate*}\item \textit{Spatio-temporal context aware}: aware of space and temporal dynamics of surrounding agents to anticipate possible interactions and avoid collision, \item \textit{Control-aware}: compliant to external and internal constraints, such as kinematic constraints, and simulation control, and \item \textit{Probabilistic}: able to anticipate multiple forecasts for any given situation, beyond those in the training data.
\end{enumerate*}

To be able to model the implicit behaviour and predict especially the sudden, unexpected changes, it is essential that these systems understand not only the spatial context but also the temporal context. This context should be identifiable, and adaptable. For instance, in urban simulations, it is important to simulate trajectories with different characteristics specific to the location and time \eg slow pedestrians in malls vs fast in busy streets, and so on. Simulations need to be able to adapt to changing environments.

In this work, we propose a generative neural network framework called CSG (Conditional Speed GAN) that takes into account the aforementioned requirements. We leverage the conditioning properties offered by conditional GANs~\cite{cgan_original_paper}, to induce temporal structure to the latent space in sequence generation system inspired by previous works~\cite{base_paper, SophiePaper, Socialbigat}. Consequently, CSG can be conditioned for controlled simulation.
CSG is trained in a self-supervised setting, on multiple contexts such as speed and agent-type in order to generate trajectories specific to those conditions, without the need for inductive bias in the form of explicit aggregation methods used extensively in previous works~\cite{base_paper,poolingref1,poolingref2,poolingref3,poolingref4, attention2, attention1}. The main contributions of this work are as follows:

\begin{enumerate}
    \item A generative system that can be conditioned on agent speed and semantic classes of agents, to simulate multimodal and realistic trajectories based on user defined control.
    \item A trajectory forecaster that uses predicted speeds from the latent space to generate conditional future moves that are socially acceptable, without special aggregation mechanisms like pooling or attention, and performs comparable to state of the art, as validated on several trajectory prediction benchmarks.
\end{enumerate}


\section{Related Work}

There is a plethora of scientific work done in the past in the field of trajectory forecasting. Based on structural assumptions~\cite{ivanovic2020multimodal}, the existing literature can broadly be classified as: \begin{enumerate*}
\item \textit{Ontological}, which are mechanics-based, such as the Cellular Automata model~\cite{handcrafted1} , Reciprocal Velocity Obstacles~(RVO) method~\cite{handcrafted2}, or the Social Forces~(SF) model~\cite{social_force_model_17} - that use dynamic systems to model the forces that affect human motion. 
For instance, SF models dynamics with newtonian controls, like, attraction towards goal, and repulsion against other agents; these methods make strong structural assumptions, and often fail to capture the intricate motion dynamics~\cite{discrete_choice_2, social_force_model_17, modelling_smooth_path_41, who_are_you_46}, and \item \textit{Phenomonological}, which are more data driven and aim to implicitly learn complex relationships and distributions. These include methods such as GPR (Gaussian Process Regression)~\cite{Gausian_Process_Regression}, Inverse Reinforcement learning~\cite{Inverse_Reinforcement_Learning}, and the more recent RNN based methods. However these methods are still restrictive, \eg, GPR suffers from long inference times~\cite{trajectron}.
RNNs have fairly recently gained traction in trajectory forecasting~\cite{Alahi_2016_CVPR1,morton2016analysis}, due to their acclaimed success in modelling long sequences, yielding an advantage in prediction accuracies over previous deterministic methods. In Social-LSTM~\cite{Alahi_2016_CVPR1}, the authors introduced a grid-based pooling method, in order to capture local intricate motion dynamics, and thus introducing spatial sense in these networks.  Inspite of their success, all these methods were limiting because of their inability to model multimodal trajectories.

\end{enumerate*}

\subsection{Generative Models:}

Generative methods, with recent advancements became the natural choice for modelling trajectories, since they offer distribution learning, rather than optimising on single best outcome. Most related works employ some kind of deep recurrent base with latent variable model, such as the Conditional Variational Autoencoder (CVAE)~\cite{schmerling2018multimodal,ivanovic2020multimodal} to explicitly encode multimodality or Generarative Adversarial Networks (GAN) to implicitly do so~ \cite{base_paper, SophiePaper, GAN_original_paper}. 
A few interesting GAN variants have been developed to tackle some of the aforementioned challenges, such as the Social GAN~\cite{base_paper}, which can produce multiple socially acceptable trajectories, and encouraged multimodal generation of by introducing a variety loss. Additionally, with the pooling module, using permutation invariant max-pooling, a kind of neighbourhood spatial embedding was introduced, that demonstrated improvement over local grid-based encoding, such as the kind used in Social-LSTM ~\cite{Alahi_2016_CVPR1}. This was improved with an attention mechanism proposed in SoPhie~\cite{SophiePaper}, which was explored by numerous following works~\cite{attention1,attention2,attention3,attention4,attention5} and improved in Social Ways~\cite{SocialWays} and Social BiGAT~\cite{Socialbigat}.

In the current state, the generative models can effectively learn distributions and forecast diverse and acceptable trajectories.
However, there still exist open questions as to how to decide which mode is best, or if the mean is good enough for changing scenarios. 
Existing methods do not tackle the problem of mode control, which is an essential characteristic needed for simulation and adaptation to different scenarios. 
Further, a key challenge is to find an ideal strategy to aggregate information in scenes with variable neighbours, and it remains unanswered whether special mechanisms like pooling or attention are really needed.

Recently graph based methods have been introduced for spatio-temporal modelling such as the Trajectron~\cite{trajectron} that takes as input the relative velocity of the neighbours to model interaction or the Trajectron++~\cite{Trajectron++}, an improved variant. In this setup, each pedestrian is denoted as a node, and two interacting pedestrians are connected with an edge. The node representations learn the trajectory sequence, while the edge representations learn the interaction sequence. 
While these methods provide state of the art results in terms of trajectory prediction metrics, such as average and final displacement error, they lack the ability of explicit control in simulation environments.

\subsection{Conditional GAN:}

The objective of generative models is to approximate the true data distribution, with which one can generate new samples similar to real data.

GANs are useful frameworks for learning from a latent distribution with only a small number of samples, yet they can suffer in regards to output mode control. 
The mode control of the network requires some additional constraints that force the network to sample from particular segments in the distribution.
Conditional GANs are an improvement over GANs that allow such kind of control, by conditioning the generation. 
As defined in~\cite{cgan_original_paper}, the objective function can be framed as a two-player minimax game between generator $G$ and discriminator $D$:
\begin{equation}
\begin{aligned}
    \min_{G} \max_{D} V(G, D)
 = \EX_{x \sim p_{data}(x)}[log(D(x|c)] + \\ \EX_{z \sim p_z(z)}[log(1-D(G(z|c)))]
\end{aligned}
\end{equation}
with $c$ as the condition and $p_z(z)$ as the noise. 
$G$ tries to model the data distribution and produce realistic ``fake" samples, whereas $D$ estimates the probability of a sample being ``real'' or ``fake" (part of the data or generated by $G$). We use these methods as the backbone in our study, where we induce further constraints on the latent vector model.

Variants of the conditional GAN have been explored in the context of trajectory prediction, conditioning on motion planning, weather effects or final goal~\cite{cganTrajectoryRobots, cGAN_Aircraft_TrajectoryPrediction, dendorfer2020goal} in order to increase prediction accuracy.
To the best of our knowledge, no previous work built on conditional GANs for simulation control nor used speed as a context vector to condition on.

\begin{subsection} {Problem Formulation} 

Trajectory prediction or forecasting is the problem of predicting the path $\textless(x^t,y^t)|t=t_{obs}+1,\ldots,T\textgreater$ that some agent (e.g., pedestrian, cyclist, or vehicle- we omit a subscript to indicate the agent here for better readability) will move along in the future given the trajectory $<(x^t,y^t)|t=0,\ldots,t_{obs}>$ that the agent moved along in the past. 

The objective of this work is to develop a deep generative system that can accurately forecast motions and trajectories for multiple agents simultaneously with user controlled speeds.

Given $(x^t, y^t)$ as the coordinates at time $t$, $L$ as the agent type and speed $S$, we seek a function $f$ to generate the next timesteps $(x^{t+1}, y^{t+1})$ as follows:
\begin{equation}
    (x^{t+1}, y^{t+1}) = f(x^{t}, y^{t}| S, L),
\end{equation}
where the generation of future timesteps is conditioned on speed $S$ and agent type $L$.
While the agent type remains constant over time, speed may vary per timestep.
In simulation environments, speed $S$ is a user-controlled variable, while in prediction environments, the speed of future timesteps is typically unknown.
In order to be able to condition on the speed of the whole timeframe, including future speeds of the yet to be generated trajectories, an estimate $\hat S$, learned from the data can be used.

\end{subsection}


\section{Methodology}

\begin{figure*}[tp]
	\centering
	\includegraphics[width=1\linewidth]{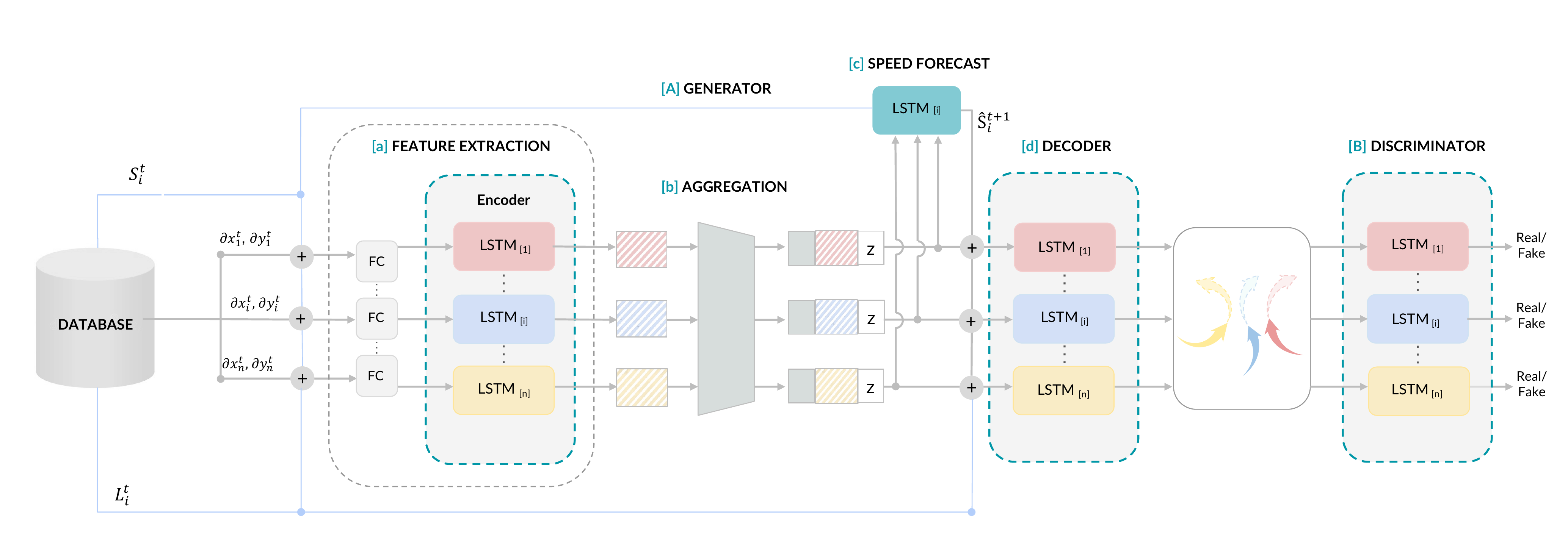}
	\caption{Overview of the CSG approach: the pipeline comprises of two main blocks: A) Generator Block, comprising of the following sub-modules: (a) Feature Extraction, that encodes the relative positions and speeds of each agent with LSTMs, (b) Aggregation, that jointly reasons multi agent interactions, (c) Speed forecast, that predicts the next timestep speed, (d) Decoder, that conditions on the next timestep speed, agent label and the agent-wise trajectory embedding to forecast next timesteps, and, the B) Discriminator Block, that classifies the generated outputs as ``real" or ``fake", specific to the conditions.}

	\label{Fig:1}
\end{figure*}

This section describes the components of the proposed Conditional Speed GAN (CSG) model. 
CSG consists of two main blocks ~(cf.~\Cref{Fig:1}): the Generator block (G) and the Discriminator block (D). G is comprised of: a) Feature Extraction module, that encodes the motion patterns of agents, b) Speed Forecasting module, which predicts the speed for the next move, c) Aggregation module, that jointly learns the agent-agent interactions, d) Decoder, that generates or forecasts trajectories conditioned on the latent space, speed and the agent label. D is composed of an LSTM encoder module encourages more realistic generation, specific to the conditions, by classifying them as ``real'' or ``fake''.

\subsection{Preprocessing:} 
We first calculate the relative positions, for translational invariance, as the difference to the previous timeframe $\delta x_i^t = x_i^t-x_i^{t-1}, \delta y_i^t = y_i^t - y_i^{t-1}$ with $\delta x_i^0 = \delta y_i^0 = 0$ from the observed trajectory for each agent $i$.
Even though the internal computation is based on relative positions $(\delta x_i^t,\delta y_i^t)$, we still use $(x_i^t,y_i^t)$ throughout the paper to ease readability. 
We calculate the speed labels based on Euclidean distance between every two consecutive timeframes from the dataset and scale them in the range (0,1). 
For the second condition, \ie, \textit{agent type}, we assign nominal labels and one-hot encode them. 

\subsection{Feature Extraction}
To extract features from past trajectory of all agents in a scene, we perform the following steps:
We concatenate the relative positions $(x_i^t,y_i^t)$ of each agent $i$ with their derived speeds $S_i^t$  and agent-labels $L_i^t$.

Next, we embed this vector to a fixed length vector, $e_i^t$, using a single layer fully connected (FC) network, expressed as:

\begin{equation}
 e_i^t =  \alpha_{e}((x^{t}_i, y^{t}_i) \oplus S_i^t \oplus L^{t}_i; W_{\alpha_e}),
\end{equation}
where, $\alpha_{e}$ is the embedding function, and $W_{\alpha_e}$ denote the embedding weights.

\subsubsection{Encoder:} 
In order to capture the temporal dependencies of all states of an agent $i$, we pass the fixed length embeddings as input to the encoder LSTM, with the following recurrence for each agent:

\begin{equation}
 h^t_{ei} =  LSTM_{enc}(e_i^t, h_{ei}^{t-1}; W_{enc}),
\end{equation}

where the hidden state is initialised with zeros, $e_i^t$ is the input embedding, and $W_{enc}$ are the shared weights among all agents in a scene.

\subsection{Aggregation Methods}

To jointly reason across agents in space, and their interaction, we employ aggregation mechanisms used widely in previous research works~\cite{base_paper, SophiePaper, Social_attention, Socialbigat}. We use the social pooling from~\cite{base_paper}, attention similar to~\cite{SophiePaper} and a simple concatenation of hidden states of $N$ nearest neighbours. The aggregation vector is computed using one of the three mechanisms per agent, and concatenated to its latent space. 
\subsubsection{Pooling}
\label{Sec: pm}
Similar to~\cite{base_paper}, we consider the positions of each agent, relative to all other agents in the scene, and pass it through an embedding layer, followed by a symmetric function.

Let $r_i^t$ be the vector with relative position of an agent $i$ to all other agents in the scene, the social features are calculated as:

\begin{equation}
f_{i}^t = \alpha_{p}{(r^t_{i}; W_{\alpha _p})},
\end{equation}

where $W_{\alpha_p}$ denote the embedding weight. The social features are concatenated with the hidden states $h_{ei}^t$ and passed through a multi-layer FC network followed by max-pooling to obtain the final pooling vectors as:
\begin{equation}
a^t_{i} = \gamma_p(h^t_{ei} \oplus f^t_{i};W_{\gamma_p}),
\end{equation}
with $W_{\gamma_p}$ as the weights of the FC network.

\subsubsection{Attention}
\label{Sec: att}
We implement a soft-attention mechanism similar to~\cite{SophiePaper}, with the difference that we compute attention only on \textit{N} nearest agents for each agent in the scene. The nearest pedestrians are sorted based on the euclidean distance between them. We compute the social features, and pass them to the attention module, with the respective hidden states from the encoder, as:

\begin{equation}
\begin{gathered}
f_{i}^t = \alpha_a(r^t_{i}; W_{\alpha_a}), \\
a^t_{i} = Attn_{so}(h^t_{ei} \oplus f^t_{i};W_{so}),
\end{gathered}
\end{equation}

where $Attn_{so}$ is the soft attention with $W_{so}$ weights.

\subsubsection{Concatenation}
\label{Sec: concat}
For each agent $i$, we calculate $N$ nearest neighbours and concat their final hidden states. The concatenated hidden states are passed through a FC network that learns the nearby agents interaction, as:

\begin{equation}
\begin{gathered}
a^t_{i} = \gamma_c(h_i^t \oplus [h^t_{en}
|\forall n \in N];W_{\gamma_c}),
\end{gathered}
\end{equation}

where $h^t_{ei}$ and $h^t_{en}$ refer to the final encoder hidden states of the current agent and $N$ nearest agents respectively,

Finally, we concatenate and embed the final hidden states of the encoder LSTMs $h^t_{ei}$ along with the respective aggregation function $a^t_{i}$ to a compressed size vector using a multi-layer FC network and add gaussian noise $z$ to induce stochasticity. 

\begin{equation}
\begin{gathered}
    h^{t}_{i} =  \gamma(h^t_{ei} \oplus a^t_{i}, W_{\gamma}) \oplus z,
\end{gathered}
\end{equation}

where $\gamma$ denotes the multi-layer FC embedding function with ReLU non-linearity , and embedding weights $W_{\gamma}$.

We treat these vectors as latent spaces to sample from for conditional generation in the following stages.
\subsection{Speed Forecasting:}

In order to forecast the future speeds for each agent in prediction environments, we use a module comprised of LSTMs. 
We initialise the hidden states of the speed forecaster $h_{si}^t$ with the latent vectors $h_i^t$.
The input is the current timestep speed $S_i^t$ and the future speed estimate $\hat S_i^{t+1}$ is calculated by passing the hidden state through a FC network with sigmoid activation in the following way:
\begin{equation}
\begin{gathered}
    h^{t}_{si} =  LSTM_{sp}(S^{t}_{i}, h^{t-1}_{si}; W_{sp}), \\ 
    \hat{S}^{t+1}_i = \gamma_{sp}(h^{t}_{si};W_{\gamma_{sp}}),
\end{gathered}
\end{equation}
The forecasting module is trained simultaneously with the other components, using ground truth $S_i^{t+1}$ as feedback signal.

\subsection{Decoder:}

As we want the decoder to maintain the characteristics of the past sequence, we initialise its hidden state $h_{di}^t$ with $h_i^t$, and input the embedded vector of relative positions with the conditions for control during training and simulation as: 
\begin{equation}
\begin{gathered}
     d^t_i =  \alpha_d((x^{t}_i, y^{t}_i) \oplus S^{t+1}_i \oplus L_i; W_{\alpha_d}).
\end{gathered}
\end{equation}
In prediction environments, we replace $S_i^{t+1}$ with the estimate $\hat S_i^{t+1}$ from the forecasting module. 
The hidden state of the LSTM is fed through a FC network that outputs the predicted relative position of each agent:
\begin{equation}
\begin{gathered}
    h^{t}_{di} =  LSTM_{dec}(d^t_i, h^{t-1}_{di}; W_{dec}), \\
    (\hat x_i^{t+1}, \hat y_i^{t+1}) = \gamma_d(h^t_{di},W_{\gamma_d}),
\end{gathered}
\end{equation}
where $W_{dec}$ are the LSTM weights and $W_{\gamma_d}$ are the weights of the FC network.

\subsection{Discriminator}
We use an LSTM encoder block as the Discriminator, which is conditioned on the agent type and speeds to encourage the Generator to not only generate more realistic and socially acceptable trajectories but also to conform to the given conditions. 
The real input to D can be formulated as:
\begin{equation}
  O_i = \textless(x_i^t, y_i^t),S_i^t,L_i|t=0,\ldots,T\textgreater,
\end{equation}
including the observed ($t=0,\ldots,t_{obs}$) and future ground truth ($t=t_{obs}+1,\ldots,T$) relative positions. The fake input can be formulated as:
\begin{equation}
\begin{aligned}
  \hat O_i = &  \textless(x_i^t, y_i^t),S_i^t,L_i|t=0,\ldots,t_{obs}\textgreater \\ 
                & \oplus \textless(\hat x_i^t,\hat y_i^t),S_i^t,L_i|t=t_{obs}+1,\ldots,T\textgreater,
\end{aligned}
\end{equation}
including the observed and predicted relative positions.

The discriminator equation can be framed as:
\begin{equation}
\begin{gathered}
 h^t_{dsi} =  LSTM_{dsi}(\alpha_{di}(o^t_{i};W_{\alpha_{di}}), h^{t-1}_{dsi}; W_{dsi}), 
\end{gathered}
\end{equation}
where $\alpha_{di}$ is the embedding function with corresponding weights $W_{\alpha_{di}}$, $W_{dsi}$ are the LSTM weights. $o^t_i$ is the input element from the real or fake sequence $O_i$ or $\hat O_i$.
The real or fake classification scores are calculated by applying a multi-layer FC network with ReLu activations on the final hidden state of the LSTMs, as:  
\begin{equation}
    \hat C_{i} = \gamma_{di}(h^t_{dsi}; W_{\gamma_{di}}),
\end{equation}

\subsection{Losses}
In addition to optimising the GAN minimax game, we apply the L2 loss on the generated trajectories, and L1 loss for the speed forecasting module.
The network is trained by minimising the following losses, taking turns:\\
The Discriminator loss is framed as:
\begin{equation}
    \ell_{D}(\hat{C_i}, C_i) = - C_i \log (\hat C_i) - (1- C_i) \log(1 - \hat C_i),
\end{equation}
The Generator loss together with L2 and L1 loss becomes:

\begin{equation}
\ell_{G}(\hat O_i) 
 + \ell_2((x^t_i,y^t_i),(\hat x^t_i,\hat y^t_i))  + \ell_1(S^t_i,\hat S^t_i),
\end{equation}
for $t=t_{obs}+1,\ldots,T$.
The Generator loss is the Discriminator's ability to correctly classify data generated by $G$ as ``fake'', expressed as:
\begin{equation}
    \ell_{G}(\hat O_i) = - \log(1 - \hat C_i),
\end{equation}
where $\hat C_i$ is the discriminator's classification score. 


\section{Experiments}

\subsection{Datasets:} For single agent predictions, we perform experiments on two publicly available datasets: ETH~\cite{eth_dataset} and UCY~\cite{ucy_dataset}, which contain complex pedestrian trajectories. The ETH dataset contains two scenes each with 750 different pedestrians and is split into two sets (ETH and Hotel). The UCY dataset contains two scenes with 786 people. This dataset has 3-components: ZARA-01, ZARA- 02 and UNIV. As shown in~\cite{pellegrini2009you}, these datasets also cover challenging group behaviours such as couples walking together, groups crossing each other and groups forming and dispersing in some scenes, and contain several other non-linear trajectories. In order to be able to test the model on multiple classes of agents, we utilise the Argoverse motion prediction dataset~\cite{Argoverse}. Similar to~\cite{GraphLstmV1}, we train and test on 5126 and 1678 samples respectively. The dataset consists of video segments, recorded in different cities like Miami and Pittsburgh with high quality multi agent trajectories. The labels available in the dataset are \textit{av}, for autonomous vehicles, \textit{agent} for other vehicles and \textit{other} includes other agents present in the scene. 
We convert the real-world coordinates from the datasets to image coordinates and plot them in real-world maps, so as to qualitatively evaluate the predictions. All plots are best viewed in colour.
\subsection{Metrics}
We compare our work in regards to the benchmark metrics followed extensively by previous works~\cite{base_paper, SophiePaper, SocialWays}: \begin{enumerate}
\item \textit{Final Displacement Error (FDE)}, which computes the euclidean distance between the final points of the ground truth and the predicted final position, and, \item \textit{Average Displacement Error (ADE)}, which averages the distances between the ground truth and predicted output across all timesteps.\end{enumerate}
We generate K samples per prediction step and report the distance metrics for the best out of the K samples.
In addition, we report average percentage of collisions per frame as a measure to evaluate the quality of generated predictions in terms of collision avoidance.
If two or more pedestrians are closer than an euclidean distance of 0.10m, we consider it as a collision.

\subsection{Simulation}
\subsubsection{Speed Extrapolation}

\begin{figure}
    \includegraphics[width = 1\linewidth]{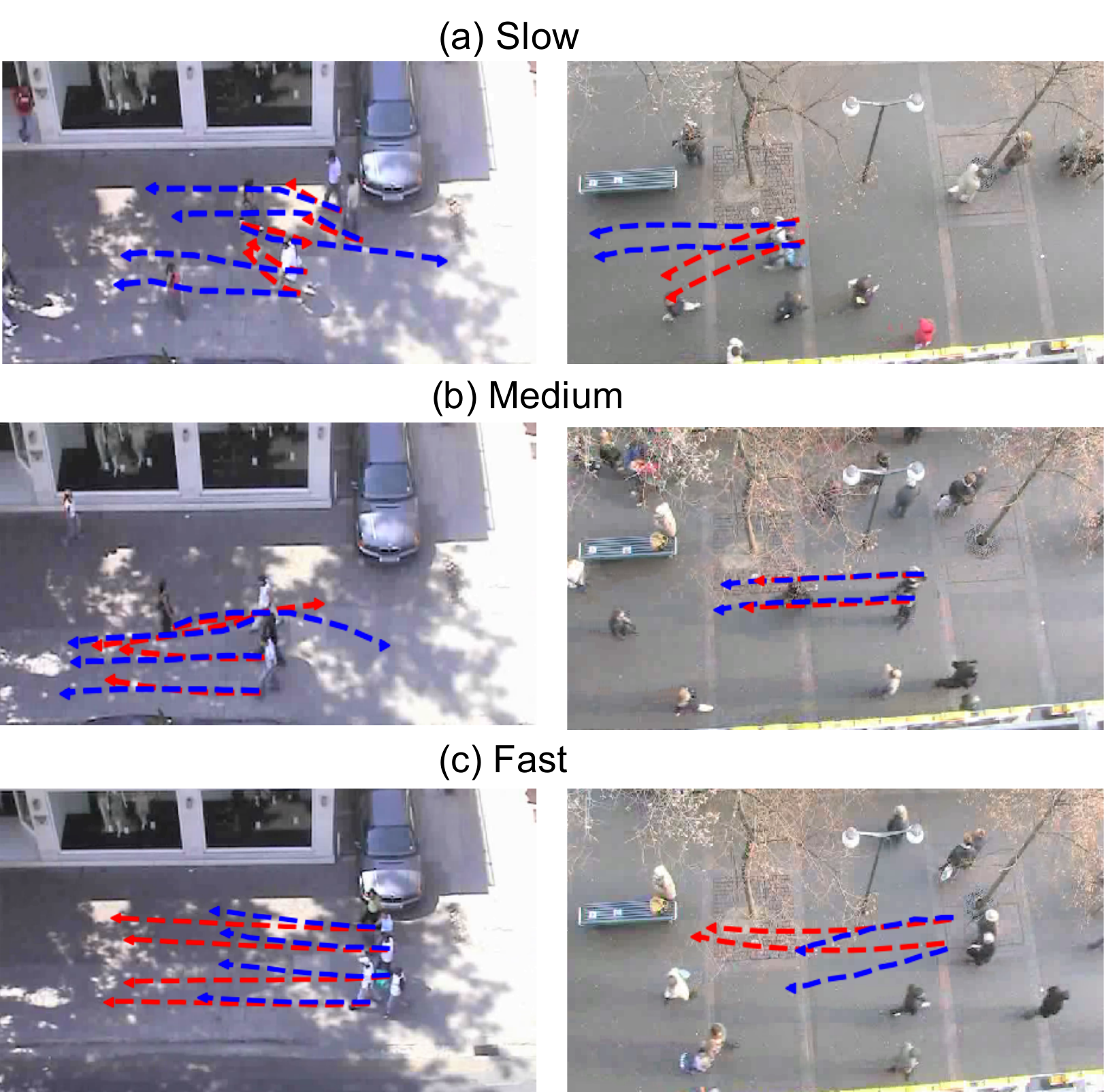}
    \caption{(a) Pedestrians from fast (on left) and medium folds (on right) simulated at slow speeds. (b) Pedestrians from fast fold simulated at medium speeds, and (c) Pedestrians from slow (on left) and medium folds (on right) simulated at fast speeds. Ground truth values are marked in blue. The network extrapolates to unseen speed contexts.}
    \label{fig: 3}
\end{figure}

We split the data into three folds according to derived speeds \ie slow, medium and fast with 0.33, 0.66 and 1 as the thresholds respectively. Using CSG with concatenation as the aggregation mechanism, we train on two folds at a time and simulate the agents in the test set of these folds with controlled speeds from the fold left out.
We observe controllability in all three segments, indicating the ability to extrapolate to unseen contexts (cf.~\Cref{fig: 3}).  In \Cref{fig: 3}(a), pedestrians from fast (on left) and medium (on right) folds are simulated at slow speeds. We clearly observe the pedestrians adapt in a meaningful way, traversing lesser distance compared to the ground truth. In \Cref{fig: 3}(b) and (c), similarly, we simulate at the medium and fast speeds unseen in the training set. 

We observe, regardless of the properties present in the training set, the network is able to extrapolate the contextual features to a certain degree, indicating some distributional changes due to localised causal interventions. In addition, to evaluate if social constraints are met, we compute the average percentage of collisions in each frame of the simulation and compare with the ground truth (cf.~\Cref{Tab: 6}). It is to be noted that collisions in the fast fold is zero, due to limited and sparse pedestrians in the ground truth of that split.
\begin{table}[ht]
    \centering
    \small
    \caption {Average percent collisions per frame for Slow, Medium and Fast folds.}
    \begin{tabular}{llll}
        \toprule
        \multirow{1}{*}{Method} & {Slow} & {Medium}& {Fast} \\
        \midrule
 GT & 0.0128 & 0.0042 & 0.0 \\
 \midrule
 CSG-C & 0.1557 & 0.2773 & 0.1310 \\
        \bottomrule
    \end{tabular}
\label{Tab: 6}
\end{table}
\subsubsection{Multimodal and socially aware}
 We demonstrate that CSG can generate diverse and socially acceptable trajectories, with simulation control.

\begin{figure*}
    \centering
    \includegraphics[width= \linewidth]{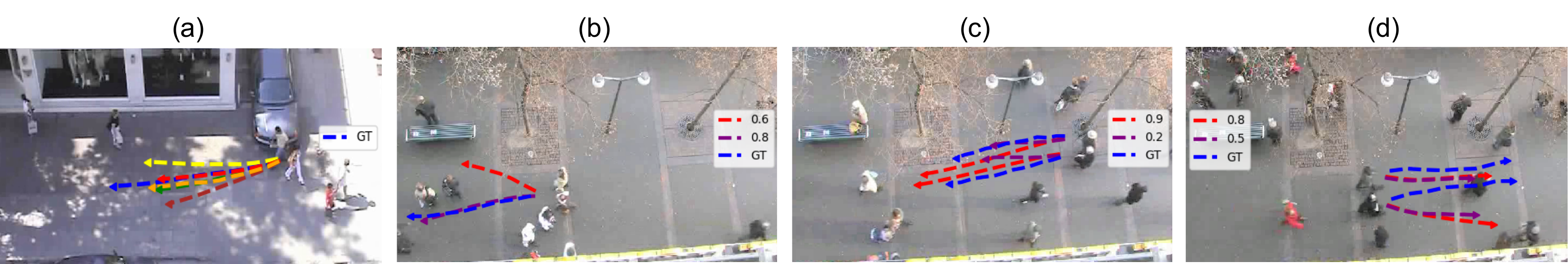}
 
    \caption{(a) A fast moving pedestrian simulated at medium speeds, with K = 5, shows a diverse selection of paths (Multimodality). (b) Two pedestrians simulated at different speeds walk around stationary people (Collision avoidance). (c) Two pedestrians walking together, simulated at fast and slow speeds, find corresponding paths in order to walk together (Group walking). (d) Two pedestrians walking together adjust their paths in order to circumvent the approaching pedestrian. All ground truth trajectories are marked in blue.}

    \label{fig: 2}
\end{figure*}

\Cref{fig: 2} illustrates the different speed control for agents predicted for 8 timeframes: (a) shows a fast moving pedestrian simulated at medium speeds with K=5, expressing a diverse generation for the controlled mode. \Cref{fig: 2}(b) illustrates preservation of social dynamics: two pedestrians simulated at different speeds circumvent a possible collision by walking around stationary people. \Cref{fig: 2}(c) depicts group walking behaviour with slow and fast simulations: the pedestrians continue to walk together, and adjust their paths in order to be able to do so. \Cref{fig: 2}(d) depicts another complex collision avoidance scenario: the pedestrians decide to split up and walk around the approaching pedestrian, when simulated at fast speeds.

\subsection{Effect of Aggregation method}

\begin{figure}
    \centering
    \includegraphics[width= 1.0\linewidth]{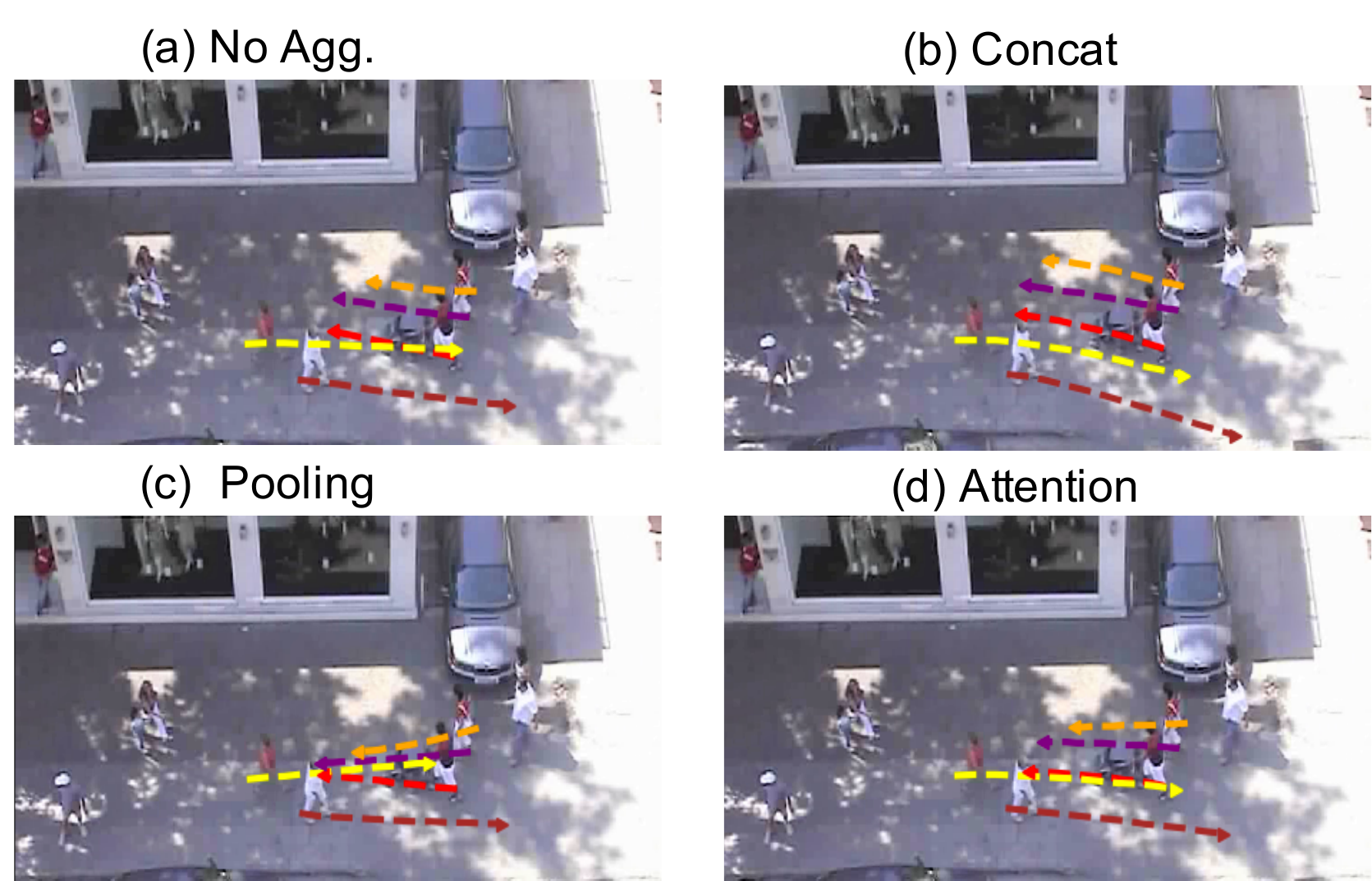}
    \caption{Effect of aggregation methods. (a) No aggregation can sometimes result in avoidable collisions. (b) Trajectories with concat as aggregation show a smoother detour around approaching pedestrians, indicating better preservation of natural dynamics compared to the pooling (c) and attention (d) variants. }
    \label{fig: 4}
\end{figure}

We evaluate the performance of our method with different aggregation strategies, one at a time, keeping all other factors constant.  CSG, CSG-P, CSG-C and CSG-A refer to our method without aggregation, with pooling, with concatenation and with attention respectively. We observe~(cf.~\Cref{Tab: 3}) that the concatenation strategy consistently outperforms all other, followed by the attention and max-pooling methods, in that order. CSG performs slightly worse in terms of collision avoidance compared to the variants with aggregation. In \Cref{fig: 4}, CSG-C appears to preserve social dynamics quite well, by generating a relatively more curved trajectory so as to avoid collision in a complex scenario. In terms of final displacement metrics~(cf.~\Cref{Tab: 4}), we observe no significant difference between either variant. CSG, without aggregation appears to replicate the ground truth trajectories the best. Regardless of the choice, CSG reduces collisions compared to SGAN, indicating that the speed forecasting module might yield some natural structure in latent space.

\subsection{Trajectory Prediction}
\subsubsection{Single agent type (pedestrian)}

\begin{table*}[ht]
    \centering
    \small
    \caption {A comparison of GAN based methods on ADE/FDE scores for 12 predicted timesteps (4.8 seconds) with K=20. For CSG, we report the metrics with the mean and variance for 20 runs. Lower is better, and best is in bold.}
    
    \begin{tabular}{llllllllll}
        \toprule
        \multirow{1}{*}{Dataset}  & {SGAN~\cite{base_paper}} & {SoPhie~\cite{SophiePaper}} & {S-Ways~\cite{SocialWays}} &  {S-BIGAT~\cite{Socialbigat}} &
        {CGNS~\cite{li2019conditional}} &
        {GoalGAN~\cite{dendorfer2020goal}} & {CSG (Ours)} \\
        \midrule
ETH  & 0.87/1.62 & 0.70/1.43 & \textbf{0.39/0.64} & 0.69/1.29 & 0.62/1.40 & 0.59/1.18 &
 \pbox{20cm}{0.81 $\pm$ 0.02/ \\ 1.50 $\pm$ 0.03} 
 \\
 \midrule
 HOTEL & 0.67/1.37 & 0.76/1.67 & 0.39/0.66 & 0.49/1.01 & 0.70/0.93 & \textbf{0.19/0.35} & \pbox{20cm}{0.36 $\pm$ 0.01/ \\ 0.65 $\pm$ 0.02}
 \\
 \midrule
 UNIV  & 0.76/1.52 & 0.54/1.24 & 0.55/1.31 & 0.55/1.32 & \textbf{0.48/1.22} & 0.60/1.19 & 
 \pbox{20cm}{0.54 $\pm$ 0.01/ \\ 1.16 $\pm$ 0.01}
 \\
 \midrule
 ZARA1 & 0.35/0.68 & \textbf{0.30/0.63} & 0.44/0.64 &  \textbf{0.30/0.62} & 0.32/0.59 & 0.43/0.87 &
 \pbox{20cm}{0.36 $\pm$ 0.02/ \\ 0.76 $\pm$ 0.01}
 \\
 \midrule
 ZARA2 & 0.42/0.84 & 0.38/0.78 & 0.51/0.92 & 0.36/0.75 & 0.35/0.71 & 0.32/0.65 & 
 \textbf{\pbox{20cm}{0.28 $\pm$ 0.01/ \\ 0.57 $\pm$ 0.02}}
 \\
 \midrule
 AVG  & 0.61/1.21 & 0.54/1.15 & 0.46/0.83 & 0.48/1.00 & 0.49/0.71 & \textbf{0.43/0.85} & 
 0.47/0.93
 \\ 

        \bottomrule
    \end{tabular}
\label{Tab: 1}
\end{table*}
\begin{table*}[ht]
    \small
    \centering
    \caption {ADE/FDE scores on Argoverse Dataset. We report our score as an average of 20 runs}
    \begin{tabular}{llllll}
        \toprule
        \multirow{1}{*}{Dataset Name} & {CS-LSTM~\cite{CS-LSTM}} & {TraPHic~\cite{Traphic}} & {SGAN~\cite{base_paper}} & {Graph-LSTM~\cite{GraphLstm2}} & {CSG (Ours)} \\

        \midrule
Argoverse & 1.050/ 3.085 & 1.039/ 3.079 & 3.610/ 5.390 & \textbf{0.99/ 1.87} & 1.39 $\pm$ 0.02/2.95 $\pm$ 0.05 \\

        \bottomrule
    \end{tabular}
\label{Tab: 2}
\end{table*}

\begin{table}[ht]
    \centering
    \small
    \caption {Average percent collisions per predicted frame. A collision is detected if the distance between two pedestrians are less than 0.10m. Lower is better, and best is in bold.
}
    \begin{tabular}{llllll}
        \toprule
        \multirow{1}{*}{Dataset}  & {SGAN}& {CSG} & {CSG-P} & {CSG-A} & {CSG-C} \\
        \midrule
 ETH  & 0.2237 & 0.3167 & 0.2603 & 0.2373~ & \textbf{0.1881~} \\
 \midrule
 HOTEL  & 0.2507 & 0.2143~ & 0.1773~ & 0.2177~ & \textbf{0.0917~}   \\
 \midrule
 UNIV  & \textbf{0.5237} & 0.5338 & 0.6064 & 0.6425~ & 0.6025~\\
 \midrule
 ZARA1  & 0.1103 & 0.0464 & 0.0660 & 0.0680 & \textbf{0.0328} \\
 \midrule
 ZARA2  & 0.5592 & 0.2184 & 0.2768 & 0.2258 & \textbf{0.1988}\\
 \midrule
 AVG  & 0.3335 & 0.2659& 0.2774& 0.2783 & \textbf{0.2228}\\
        \bottomrule
    \end{tabular}
\label{Tab: 3}
\end{table}

\begin{table}[ht]
    \small
    \centering
    \caption {Effect of aggregation method. ADE/FDE scores for the different CSG variants on 12 predicted timesteps~(4.8 s).}
    
    \begin{tabular}{lllll}
        \toprule
        \multirow{1}{*}{Dataset} & {CSG} & {CSG-P} & {CSG-A} & {CSG-C} \\
        \midrule
 ETH & \textbf{0.81/1.50} & 0.82/1.56 & 0.89/1.65 & 0.82/1.56 \\
 \midrule
 HOTEL & 0.36/0.65 & 0.34/0.64 & \textbf{0.33/0.59} & 0.34/0.63   \\
 \midrule
 UNIV & \textbf{0.54/1.16} & 0.58/1.18 & 0.66/1.38 & 0.62/1.31 \\
 \midrule
 ZARA1 & 0.36/0.76 & \textbf{0.35/0.72} & 0.35/0.73 & 0.37/0.76  \\
 \midrule
 ZARA2 & \textbf{0.28/0.57} & 0.30/0.63 & 0.29/0.60 & 0.31/0.65  \\
 \midrule
 AVG & \textbf{0.47/0.93} & 0.48/0.95 & 0.50/0.99 & 0.49/0.98  \\
        \bottomrule
    \end{tabular}
\label{Tab: 4}
\end{table}

We evaluate our model on the five sets of ETH and UCY data, with a hold-one-out approach (\ie training of four sets at a time and evaluating on the set left out) and compare with the following baseline methods: \\
    \textbf{SGAN}~\cite{base_paper}: GAN with a pooling module to capture the agent interactions, \\
    \textbf{SoPhie}~\cite{SophiePaper}: GAN with physical and social attention, \\
    \textbf{S-Ways}~\cite{SocialWays}: GAN with Information loss instead of the L2, \\
    \textbf{S-BIGAT}~\cite{Socialbigat}: Bicycle-GAN augmented with Graph Attention Networks~(GAT), \\
    \textbf{CGNS}~\cite{li2019conditional}: CGAN with variational divergence minimization, and \\
    \textbf{Goal-GAN}~\cite{dendorfer2020goal}: GAN that predicts the final goal position and conditions its generation on it.

\Cref{Tab: 1} depicts the final metrics for 12 predicted timesteps (4.8 seconds). Similar to other methods~\cite{base_paper, Socialbigat}, we generate K=20 samples for a given input trajectory, and report errors for the best one. On a quantitative comparison with other GAN models (cf.~\cref{Tab: 1}), we observe that our model outperforms SGAN, Sophie, S-Bigat and CGNS in HOTEL and ZARA2, while performs competitively in other datasets. S-Ways, SoPhie and Social-Bigat perform on par with CSG in the UNIV dataset however CSG constantly outperforms S-Ways in ZARA1, ZARA2 and HOTEL datasets. In comparison with GoalGAN, our model performs better in UNIV, ZARA1 and ZARA2 datasets while GoalGAN performs relatively better in ETH and HOTEL datasets. Overall, CSG performs best on ZARA2 and on average is comparable to the state-of-the-art GAN based methods.

\subsubsection{Multi agent type (Argoverse Dataset)}

With respect to multi agent problem, we compare our model with:\\
    \textbf{CS-LSTM}~\cite{CS-LSTM}: Combination of CNN network with LSTM architecture \\
    \textbf{TraPHic}~\cite{Traphic}: Combination of CNN LSTM networks integrated with spatial attention pooling \\
    \textbf{SGAN}~\cite{base_paper}: GAN network with max pooling approach to predict future human trajectories \\
    \textbf{Graph-LSTM}~\cite{GraphLstm2}: Graph convolution LSTM network using dynamic weighted traffic-graphs that predicts future trajectories and road-agent behavior.

We utilise the first 2 seconds as observed input to predict the next 3 seconds and report the metrics in \Cref{Tab: 2}. We observe that our model outperforms SGAN~(cf.~\cref{Tab: 2}) by a large margin and performs better than TraPHic, CS-LSTM in terms of FDE but doesn't perform as well when compared with ADE. Graph-LSTM performs overall the best. However, CSG can explicitly control generation of heterogeneous agents and with user defined speeds.


\section{Conclusion and Future Work}
We present a method for generation and controlled simulation of diverse multi agent trajectories in realistic scenarios. We show that our method can be used to explicitly condition generation for greater control and ability to adapt context. Further, we demonstrate with our experiments the efficacy of the model in forecasting mid-range sequences~(5 seconds) with an edge over most existing GAN based variants. It may be that most models are optimised to reduce the overall distance metrics, but not collisions. The models are expected to learn the notion of collision avoidance implicitly. By focussing explicitly on relative velocity predictions, we obtain more domain knowledge driven control over the design of the interaction order. Further, we observe that a simple concatenation of the final hidden state vectors of $N$ neighbours is good enough strategy for aggregating information across agents in a scene. While this approach relatively simple, it is efficient, and removes the need to design complex mechanisms. Finally, we acknowledge there is room for improvement. This method could be extended by learning context vectors of variation automatically and interpreting them. Additionally, it might be useful to explore techniques to optimise on social dynamics such as collision avoidance and to condition with static scene information to improve interactions in space.


{\small
\bibliographystyle{unsrt}
\bibliographystyle{ieee_fullname}
\bibliography{CSG}
}

\end{document}